\documentclass[11pt,a4paper]{article}
\usepackage[hyperref]{naaclhlt2019}
\usepackage{times}
\usepackage{latexsym}
\usepackage{url}
\usepackage{booktabs} % For formal tables
\usepackage{amsmath}
\usepackage{algorithm}
\usepackage[noend]{algpseudocode}
\usepackage{xcolor}
\usepackage{multirow}
\usepackage{graphicx} 
\graphicspath{ {Figure/} }

\aclfinalcopy % Uncomment this line for the final submission
\title{Context-Sensitive Malicious Spelling Error Correction}

\author{Hongyu Gong, Yuchen Li, Suma Bhat, Pramod Viswanath \\
  University of Illinois at Urbana-Champaign \\
  {\tt \{hgong6, li215, spbhat2, pramodv\}@illinois.edu} \\}
 
\begin{document}
\maketitle
\begin{abstract}
  Misspelled words of the malicious kind  work by changing specific keywords and are intended to thwart existing automated applications for cyber-environment control  such as harassing content detection on the Internet and email spam detection. In this paper, we focus on malicious spelling correction, which requires an approach that relies on the  context and the surface forms of targeted keywords. In the context of two applications--profanity detection and email spam detection--we show that malicious misspellings seriously degrade their performance. We then propose a context-sensitive approach  for malicious spelling correction using word embeddings and demonstrate its superior performance compared to state-of-the-art spell checkers.
\end{abstract}

\section{Introduction}

% motivation for spelling error correction

Automatic spelling correction has been an important  natural language processing component of any input text interface of today \cite{jurafsky2014speech}.  While spelling errors in common writing could be regarded as
innocuous omissions,  affecting opinions about the writer's competence at worst, those of the malicious kind can potentially be highly offensive, since they are intended to deceive an automated mechanism--be it a spam filter for emails or a profanity detector for social media. 
For those applications that rely on automatic detection of specific keywords to enable real-time control of the cyber-environment, detecting and correcting spelling errors is of primary importance. 
%Social media provide open platforms for speech, but the lack of efficient supervision might cause problems such as abusive content. Automatic detection of online toxic content has been studied to  enable real-time control of the cyber-environment. 

Perspective \cite{perspective}, is one such application that helps reduce abusive content online by detecting profane language. It has been pointed out that Perspective's otherwise  good performance in detecting toxic comments  is not robust to spelling errors \cite{hosseini2017deceiving}. 
Consider the sentence, ``My thoughts are that people should stop being stupid.'' which was assigned a perceived toxicity score of 86\% by Perspective.  When the word ``stupid'' was misspelled as ``stup*d'' in the same sentence, its toxicity score reduces to $8\%$. This marked reduction induced by the deceptive spelling error on the keyword  `stupid'  reflects the importance of accurate spelling correction for optimal toxicity detection. Had the spelling error been detected and corrected \textit{before} scoring for toxicity, the score would not have lowered. 

Likewise, deliberate typographic errors are common in phishing scams and spam emails. For instance, ``Please pya any fees you may owe to our company'' where \emph{pya} is clearly a spelling error,  included to deceive spam filters that are sensitive to keywords.  In both these instances, spelling correction as a preprocessing step of the messages are critical for a robust performance of the  target system. 
Spelling correction in the preprocessing stage in malicious settings constitutes the focus of this study.

In this paper, we empirically demonstrate the effect of spelling errors in a malicious setting by adding synthetic misspellings to sensitive words in the context of two applications--profanity detection and spam detection. We propose an \textit{unsupervised} embedding-based algorithm to correct the targeted misspelled words. Earlier approaches to spelling correction primarily depend on the edit distance to find  words morphologically similar to corrections. More recently, spell checkers have been improved with the addition of contextual information (e.g., n-gram modeling in  \cite{wint2018non}), often with intensive computation and memory requirements (to obtain and store N-gram statistics).  A recently studied neural network-based spell checker  learns the misspelling pattern from annotated train data in a supervised way, and was found to be sensitive to data domains and dependent on human annotations \cite{ghosh2017neural}. 
Our approach to make  the correction procedure context-aware involves harnessing the geometric properties of word embeddings. It is light-weight in comparison, unsupervised, and can be adapted to a variety of data domains including Twitter data and spam data, since domain information can be well captured by tuning embeddings on domain-specific data \cite{taghipour2015semi}. % \textcolor{red}{citation}. %Some research works such as neural network based spell checker also learn the misspelling pattern from annotated train data in a supervised way, and this method is sensitive to data domains and dependent on human annotations \cite{ghosh2017neural}. 

We first demonstrate the effect of spelling errors in a malicious setting in the context of two applications--profanity detection and spam detection. 
Then we propose an unsupervised  algorithm to correct the targeted spelling errors. Besides performing well on the correction of synthetic spelling errors, the algorithm also performs well on  spell checking of real data, and shows that it can improve hate speech detection using Twitter data. %check
%Our spelling correction system is available in the supplementary material.

%Deliberate typos are also found in phishing scams. An example is ``Please pya any fees you may owe to our company'' where \emph{pya} is clearly a typo. These typos are included to deceive spam filters which are sensitive to keywords.  Spelling correction as the preprocessing of input emails is hence important for a robust scam blocking system.

\section{Related Work}
%Automatic spelling correction is an important problem in natural language processing with a long research track \cite{gubanov2014improved,hassan2013social,farra2014generalized}, including statistical models  \cite{jurafsky2014speech}, and those using contexts \cite{ambercentral,baziotis-pelekis-doulkeridis:2017:SemEval2}.

\textbf{Spelling correction}. Automatic spelling correction in non-malicious settings has  had a long research track as an important component of text processing and normalization. Early works  have used edit distance to find morphologically similar corrections \cite{ristad1998learning}, noisy channel model for misspellings \cite{jurafsky2014speech}, and iterative search to improve corrections of distant spelling errors \cite{gubanov2014improved}. Word contexts have been shown to be improve the robustness of spell checkers with n-gram language model as one approach to incorporate contextual information  \cite{hassan2013social, farra2014generalized}. Other ways of incorporating contextual information  include n-gram statistics capturing the cohesiveness of a candidate word with the given context \cite{wint2018non}. 

%The proposed system is based on constructing a lattice from possible nor- malization candidates and finding the best normal- ization sequence according to an n-gram language model using a Viterbi decoder.
\noindent\textbf{Effect of malicious misspellings}. Misspellings are commonly seen in online social platforms such as Twitter and Facebook, and recent studies have drawn attention to the fact that online users deliberately introduce spelling errors to thwart bullying detection \cite{power2018detecting} or to challenge to moderators of online communities \cite{papegnies2017impact}. This is because  simple ways of filtering terms of profanity  are rendered inadequate in the presence of spelling errors. Likewise, email spam filters  are often obfuscated by misspelled sensitive words  \cite{zhong2014deobfuscation} because word-based spam filters tend to make false  decisions in the presence of misspellings \cite{saxena2015spamizer}, and so are word-based cyberbullying detection systems \cite{agrawal2018deep}. Misspelling is also seen in web attacks where a phishing site has a domain name as a misspelling of a legitimate website \cite{howard2008web}.

\noindent\textbf{Approaches to correct malicious misspellings}. Given the malicious intent of misspellings in the context of the Internet, recent studies have proposed correction strategies specifically for malicious misspellings. Levenshtein edit distance is commonly used in spell checking of detection systems. Spam filters \cite{zhong2014deobfuscation} and cyberbullying detection systems \cite{wint2017spell} rely on the idea of edit distance to recognize camouflaged words by  capturing the pattern of misspellings. For example, \cite{rojas2013revealing} studied an edit distance function designed to reveal instances where spammers had interspersed black-listed words with non-alphabetic symbols. Lexical and context statistics, such as  word frequency, have also been used to make corrections in social texts
%, where  words with the highest likelihood were selected as corrections 
\cite{baziotis-pelekis-doulkeridis:2017:SemEval2,jurafskyspeech}. {Existing approaches have the problem of domain specificity, since their lexical statistics are obtained from a certain domain which might differ from the domain of target applications.}

%\cite{ambercentral}
%\cite{baziotis-pelekis-doulkeridis:2017:SemEval2}
Our study considers the spelling correction in a malicious setting where errors are not random, but are carefully  introduced. 
%For the purpose of the problem in this study, we consider a specific case of spelling correction--one that happens in a malicious setting--where the errors are carefully introduced to thwart the detector. Here, the errors are not random, but are carefully introduced. 
Our context-aware approach to spelling correction  relies on the geometry of word embeddings, {which has the advantage of efficient domain adaptation. As we will show in Section~\ref{subsec:corr_context}, the embedding can be easily tuned on a small corpus from the target domain to capture domain knowledge.}
%computation efficiency over lexical statistics based approach %\textit{red}{help me replace the following "statistical models" with something more informative} statistical models.

\section{Methods}
\label{sec:addError}
We study malicious spelling correction for three target applications--toxicity detection using the Perspective API, email spam detection, and hate speech detection on Twitter data. We work with synthetic spelling errors in the first two applications, and study the real misspellings in the third.

For the toxicity detection task, we use the Perspective dataset \cite{perspective}, which provides a set of comments collected from the Internet with human annotated scores of toxicity. Using the Perspective API we obtained the toxicity score for $2767$ comments in the  dataset.

The spam dataset consists of $960$ emails from the Ling-Spam dataset \footnote{\url{http://csmining.org/index.php/ling-spam-datasets.html}}. We randomly split the data into $700$ train emails and $260$ test emails. Both the train and test sets were balanced with respect to the positive and negative classes. The most frequent $2500$ words in the training data were selected, and we counted the occurrence of each word in the emails. The number of occurrences of these frequent words were used as a $2500$-dimension feature vector for each email. We first used these features to train a Naive Bayes classifier. It achieved an accuracy of 98\% in spam detection. 

As for the Twitter data,  a total of 16K user posts were collected from Twitter, a social communication platform and used in  used in \cite{waseem2016hateful}. Of these tweets,  $1937$ were labeled as being racist,  $3117$ were labeled as being sexist, and the rest were neither racist nor sexist. The task of hate speech detection is to classify these tweets into one of the three  categories {\textit{racist, sexist, neither}}. We randomly split the tweets into train and test data, and trained a neural-network based hate speech detection system on the training data (described later in Section 5.3).

% types of spelling error
\subsection{Characterization}
%Here we summarize the characteristics of the spelling errors by grouping them into broad categories.
Here we summarize our assumptions on the characteristics of the  malicious spelling errors. \\
%\begin{itemize}[noitemsep,topsep=0pt]
(1) Malicious errors are usually made on the sensitive keywords in order to obfuscate the true intentions and deceive detection systems that rely on keywords \cite{weinberger2009feature}. \\
(2) The error yields a word that is similar to the original word in surface form. The misspelled words would be words that humans can easily understand (thus the erroneous words still convey the intended meaning, while being in disguise).  Their similarity is reflected in the small edit distance between the erroneous and the correct word \cite{howardspelling}. The errors often involve character-level operations as shown in Table \ref{tab:errorExample} \cite{liu2011insertion}. \\
(3) With edit distance as the similarity criterion, it is often the case that the word with an error is similar to multiple valid words, making correction a challenge in real applications. For example, both \emph{stud} and \emph{stupid} can be thought of as the correct form of the erroneous word \emph{stupd}. In such cases, spelling correction relies on the context in which the word occurs.
%\end{itemize}

% introduce our method and contribution
\subsection{Effect  of malicious spelling errors}
\label{sec:adv_error}
We quantify the effect of spelling errors by showing that a simple mechanism of injecting malicious spelling errors greatly degrades the performance of toxicity detection and spam detection. Toward this, we first describe  the general mechanism to generate synthetic errors and then study their impact on toxicity and spam detection. We point out that owing to the absence of a dataset with intended malicious errors, we had to generate them synthetically. 

%\begin{figure}[htbp!]
%\centering
%\includegraphics[width=0.5\textwidth]{error_system.jpg}
%\caption{Malicious Error System}
%\label{fig:errSys}
%\end{figure}

{\bf Spelling error generation.} 
%Fig.~\ref{fig:errSys} illustrates how malicious errors are added.
We first choose the ``sensitive'' words that  the detection algorithm relies (the mechanism will be discussed separately for each application we consider). We then replace these words with their erroneous forms (which may not be real words). Given the characteristics of malicious errors mentioned in Section 3.1, our assumption is that as a result of these errors the altered words will appear similar to the original ones. As illustrated in Table \ref{tab:errorExample}, we consider four basic character-level operations to change the sensitive words:  insertion,  permutation, replacement, and removal. {In our experiments we  perform these operations on randomly picked characters of the sensitive word.} For permutation, we exchange this character with the next one.
Each operation in Table~ \ref{tab:errorExample} increases the edit distance by 1, and we generate malicious misspellings of the sensitive word that are at most  2  edit distance from the correct word.
%\textcolor{red}{how did you decide which character to operate on?}
\begin{table}[htbp!]
\centering
\caption{Common spelling errors}
\label{tab:errorExample}
\resizebox{0.45\textwidth}{!}{
\begin{tabular}{|c|c|c|c|c|c|}
\hline
Type & insertion &  permutation & replacement & removal \\ \hline
Error & idio.t &  moeny & chanse & stupd \\ \hline
Correction & idiot &  money & chance & stupid \\ \hline
\end{tabular}}
\end{table}

%\textcolor{red}{how did you pick these sensitive words?} \textcolor{red}{A: Illustrated below.}

Finally, we obtain revised sentences by replacing the sensitive words in the original sentences with their erroneous counterparts. Next, we introduce our mechanism for selecting the ``sensitive'' words for each application.

%\textbf{Begin here!!!}
\textbf{Perspective toxicity detection}.
%The Perspective dataset \textcolor{red}{citation?} provides a dataset of comments taken from the internet with a range of human annotated toxicity. Using the Perspective API we obtained the toxicity score for each  comment in the Perspective dataset. 
We select $2767$ clean comments from a total of $2969$ comments in the Perspective data, with the selection criteria  given below.
%\textbf{Select comments in Perspective dataset}: we apply the criteria below to select clean comments.
\begin{itemize}
\item The most toxic word in a comment should contain more than two characters. A word that is too short is not likely to be a meaningful toxic word, and can have too many candidate corrections in the dictionary.
\item The most toxic word in a comment should appear at least 100 times in the dataset, since rare words tend to be misspellings in online texts.
\item The most toxic word in a comment should be a content word, i.e., it should not belong in the list of function words such as %``be'', ``am'', ``are''. %, ``is'', ``was'', ``were'', ``being'', ``been'', ```can'', ``could'', ``may'', ``might'', 
``must'', ``ought'', ``shall'' and``should''. %, ``will'' and ``would''.
Function words are not toxic and so were excluded from our experiments.
\end{itemize}

%For each comment, we delete a word and note the resulting predicted toxicity score in the range of $[0,1]$. 
For each comment, its predicted toxicity score lies in the range of $[0,1]$. The higher the score, the more toxic the sentence is. For each sentence, we chose the most toxic word to be that word, without which the toxicity score reduced the most. We then added malicious errors, as described above, to this word. We note that {for $2380$ toxic comments with toxicity scores higher than 0.5,  spelling errors brought down their predicted toxicity by $32\%$.}
Section \ref{sec:exp} provides the details of the  degradation in toxicity detection.
%\textcolor{red}{Comments: degradation in performance was only mentioned for spam. need to include this for toxicity as well. We should do it in section 3.2 itself and not in section 5}

%The spam dataset consists of $960$ emails from Ling-Spam dataset \footnote{\url{http://csmining.org/index.php/ling-spam-datasets.html}}. We randomly split the data into $700$ train emails and $260$ test emails. Both train and test data have equal number of spams and non-spams. The most frequent $2500$ words in training data are selected, and we count the occurrences of each word in emails. The occurrences of these frequent words are used as a $2500$-dimension feature vector of emails. We first use these features to train a Naive Bayes classifier. It achieves 98\% accuracy in spam detection. 

\textbf{Spam detection}. A Naive Bayes spam detection model provides the likelihood of each word given the spam and non-spam classes. For  a given word, the difference between these probabilities  reflects how important that word is in spam detection. We sorted all the words in decreasing order of this difference in probabilities, and  picked the most spam-indicative words. We then added malicious errors to these words in test spam emails. For a spam filtering system which relies heavily on the counts of spam-indicative words, the errors can easily disguise spams as non-spams. This is seen in the test accuracy dropping from $98\%$ on the original test data to $72\%$ on the revised test data.

We note that for the Perspective data, we added errors to only one word per sentence, but did not have such limits for the spam data. Also, we do not limit the number of corrections during the spell check process.

\section{Spelling Correction}
\label{sec:model}
%\begin{figure*}[htb]
%\centering
%\includegraphics[width=0.9\textwidth]{correction_system.jpg}
%\caption{Correction System}
%\label{fig:corSys}
%\end{figure*}

%The spelling correction system is shown in Fig.~\ref{fig:corSys}.
We have shown that malicious errors degrade the performance of toxicity detection and spam detection. In this section, we introduce our unsupervised algorithm on non-word error correction based on the relevant context. 

Our correction method proceeds in two stages. In the first stage of {\bf candidate enumeration}, the algorithm detects the spelling error and proposes candidates that are valid words and similar in surface form to the misspelled word. In the next stage,  {\bf correction via context},  the best candidate is chosen based on its context.
%We focus on the case when there is at most one typo in each sentence. 

\subsection{Candidate Enumeration}
As in the case of non-word spelling correction \cite{jurafsky2014speech}, we check whether an erroneous word is valid or not using a vocabulary of valid words. For toxicity detection and spam detection, the vocabulary consists of words that occur more than $100$ times in English Wikipedia corpus \footnote{available at: \url{http://www.cs.upc.edu/~nlp/wikicorpus/}} %\textcolor{red}{which Wikipedia dump? is there a version for the Wikicorpus?}. 
For hate speech detection, the vocabulary consists of standard words from  Wikipedia and the list of internet slang words as detailed in Section~\ref{sec:exp_cyberbullying}, since slang words frequently occur in tweets. %\textcolor{purple}{what is the dictionary of valid words?} 
For an invalid word, we  find candidate words with the smallest Damerau-Levenshtein edit distance \cite{damerau1964technique,levenshtein1966binary}. This distance between two strings is defined as the minimum number of unit operations required to transform a string into another, where the unit operations include character addition, deletion, transposition and replacement. We note that because there are potentially multiple candidates having the smallest distance to the misspelled word, the output of this stage is a set of candidate corrected words.

\subsection{Correction via Context}
\label{subsec:corr_context}
After enumerating all possible candidates, we  choose the one that fits the best in the context.  We propose a word-embedding-based method to match the context with the candidate.

\noindent\textbf{Pretrained embedding}. %Word embedding is the low-dimension vector representation of words commonly used in downstream applications of natural language processing. In this work, we also make use of the geometric properties of candidate words and their contexts to measure their relatedness. 
%Depending on the domain,  a language is used in a specific style.
We use word2vec CBOW model to train embeddings \cite{NIPS2013_5021}.
The Perspective and Twitter data  have an informal style, while email data consists of relatively formal expressions. Word embeddings are known to be domain-specific \cite{nguyen2014employing} and naturally  domain-specific corpora are used to train word embeddings for use in a given domain. We note that in the absence of a large enough representative corpus  to train domain-specific high-quality embeddings for this study, we reconcile with word embeddings  trained on a large WikiCorpus \cite{polyglot:2013:ACL-CoNLL} to capture the general lexical semantics and further tuned on a domain-specific corpus, such as that of the Perspective dataset, the spam emails data and the Twitter data. %\textcolor{purple}{does this capture what you did? Does this mean that we used two sets of domain-specific word embeddings?}
 This step allows us to combine domain information in trained embeddings. 

%\vspace{-2pt}

\noindent\textbf{Error Correction}. 
%The representation of words as vectors 
Word vectors permit us to decide the fit of a word in a given context by considering the geometry of the context words in relation to the given word.
%Word embeddings have some nice properties which could reflect lexical semantics. 
Firstly, the embedding of the compositional semantics of phrases or sentences can be approximated by a linear combination of the embeddings of their constituent words \cite{salehi2015word}. Let $v_{w}$ be the embedding of the token $w$. Take the phrase ``lunch box'' as an example, it holds that
\begin{align*}
v_{\text{hate\_group}} \approx \alpha v_{\text{hate}} + \beta v_{\text{group}},
\end{align*}
where $\alpha$ and $\beta$ are real-valued coefficients.

This enables us to represent the context as a linear space of the component words. Another property is that semantically relevant words lie close in the embedding space. If a word fits the the context, it should be close to the context space, i.e., the normalized Euclidean distance from the word embedding to the context space should be small \cite{gong2017geometry}. We quantify the inconsistency between the word and its context by the distance between the word embedding and the span of the context words.

For a misspelled word in a sentence, all words occurring within a window of size $p$ are considered to be its context words. Let the context $T_{p}$ be the set of words within distance $p$ from $w_{0}$:
$$T_{p}=\{w_{-p}, w_{-p+1},\ldots, w_{0}, \ldots, w_{p-1}, w_{p}\},$$ 
where $w_{0}$ is the misspelled word.  
Let $C$ be the set of candidate replacements of $w_{0}$.
Let $v_{i}$ be the word embedding of context word $w_{i}$. We denote by $\text{dist}(\text{c}, T_{p})$,  the distance between a candidate $c$ and the linear span of the words in its context $T_p$, termed as the candidate-context distance  of a candidate, % $v_{c}~(c\in C)$ and the linear span of context words within distance $p$. 
defined as the normalized distance between the word embedding of the candidate, $v_c$  and its linear approximation obtained by the context vectors:
\begin{align}
\text{dist}(c, T_{p}) = \min\limits_{\{a_{i}\}}\frac{1}{\lVert v_{c} \rVert_{2}} \lVert \sum\limits_{\substack{i=-p,\\ i\not=0}}^{p}a_{i}v_{i}-v_{c}\rVert_{2}^{2}.
\end{align}
This is a quadratic minimization problem for which we can find a closed-form solution.% to $\text{dist}(c, T_{p})$.

Instead of fixing one context window size, we consider multiple window sizes and weigh the distances obtained using different window sizes. The context words that are closer to the misspelled word are more informative in candidate selection. In the sentence ``the stu*pid and stubborn administrators'', the word \emph{stubborn} suggests that the misspelled word should be a negative personality adjective, and the word \emph{administrator}  that it should be an adjective for people. Thus, the closest context word \emph{stubborn} provides more relevant information than the distant word \emph{administrator}.

We thus weigh the distance by the inverse of the context window size, i.e., the weight $\frac{1}{p}$ for the window size $p$. Suppose that $T$ is the full context, and the candidate-context distance is defined below:
\begin{align}
\text{dist}(c, T) = \sum\limits_{p=1}^{P}\frac{1}{p}\cdot\text{dist}(c, T_{p}),
\end{align}
where $P=4$ in our experiments.

%\textcolor{red}{how do you obtain the ai's --coefficient of $v_i$ in the equation for candidate context distance?}

Given the context $T$, the suggested correction $c^{*}$ is the candidate with the smallest distance to the linear space of context words, i.e., 
\begin{align}
c^{*}=\arg\min\limits_{c\in C} \text{dist}(c, T).
\end{align}

\begin{table}[tbp!]
\centering
\caption{Correction Accuracy on Perspective and Spam}
\label{tab:acc}
\resizebox{0.48\textwidth}{!}{
\begin{tabular}{|c|c|c|c|c|}
\hline
Dataset & Our system & PyEnchant & Ekphrasis & Google \\ \hline
Perspective & {\bf 0.840} & 0.476 & 0.713 & 0.323 \\ \hline
Spam & {\bf 0.786} & 0.618 & 0.639 & 0.526 \\ \hline
\end{tabular}}
\end{table}

\begin{table*}[htb!]
\centering
\caption{Correction on perspective data}
\label{tab:perspective_example}
\resizebox{\textwidth}{!}{
\begin{tabular}{|l|l|l|l|}
\hline 
original sentence & the {\bf stupid} and stubborn administrators & anti American {\bf hate} groups & you're a biased {\bf fuck} \\ \hline
revised sentence & the {\bf stu*pid} and stubborn administrators & anti American {\bf ahte} groups & you're a biased {\bf fucdk} \\ \hline
Our correction & the {\bf stupid} and stubborn administrators & anti American {\bf hate} groups & you're a biased {\bf fuck} \\ \hline
PyEnchant & the {\bf sch*pid} and stubborn administrators & anti American {\bf hate} groups & you're a biased {\bf Fuchs} \\ \hline
Ekphrasis & the {\bf stupid} and stubborn administrators & anti American {\bf ate} groups & youre a biased {\bf fuck} \\ \hline
Google & the {\bf stupid} and stubborn administrators & anti American {\bf ahte} groups & you're a biased {\bf fucdk} \\ \hline
\end{tabular}}
\end{table*}

\section{Experiments}
\label{sec:exp}
We evaluated our spelling correction approach in three settings: toxicity and spam detection with synthetic misspellings, and hate speech detection with real spelling errors. Even though some recent works using neural networks \cite{ghosh2017neural} are available, they require ground truth corrections for supervised training. For our unsupervised approach, we compare its performance with that of three strong unsupervised baselines below, which are generic off-the-shelf tools used in many NLP systems such as AbiWord \cite{abiword} and Google search engine \cite{google_search}. \\
%\begin{itemize}[noitemsep,topsep=0pt]
(1) Pyenchant: a generic spell checking library of multiple correction algorithms \cite{enchant}. We use the MySpell library in this work.\\
(2) Ekphrasis: a spelling correction and text normalization tool for texts from social networks \cite{baziotis-pelekis-doulkeridis:2017:SemEval2}.\\
(3) Google spell checker: a Google search engine-based spell check  \cite{google}. We chose it for its ability to undertake context-sensitive correction on the basis of user search history.
%\end{itemize}

\subsection{Toxicity Detection}
\label{sec:exp_toxicity}
\begin{figure}[htbp!]
\centering
\includegraphics[width=0.45\textwidth]{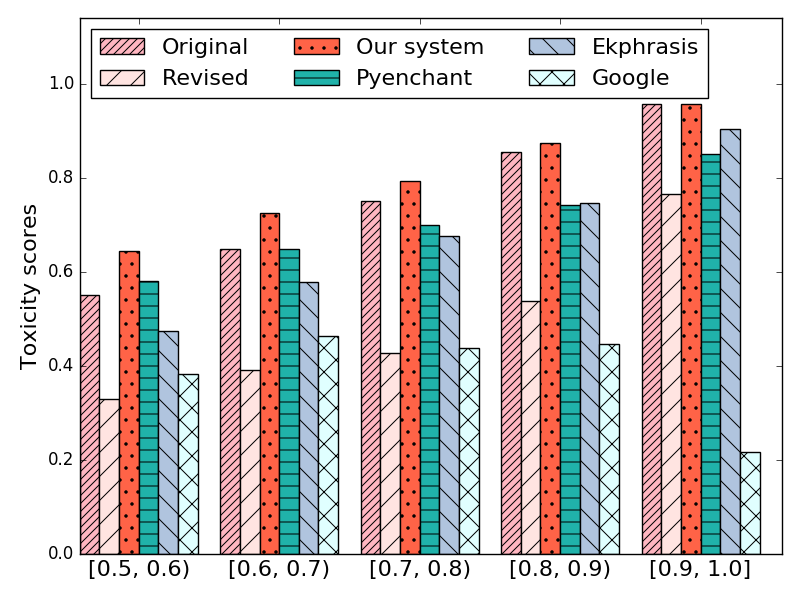}
\caption{Toxicity scores by different approaches}
\label{fig:perpective_plot}
\end{figure}

%\textcolor{red}{can you arrange the dataset description all in one place?It is better early on so section 3 has access to that information.}
%\subsubsection{Synthetic Misspellings}

We took $2380$ sentences from Perspective dataset whose toxicity scores were higher than $0.5$ and added synthetic errors maliciously to these sentences as described in Section \ref{sec:addError}. Some example Perspective sentences, revised sentences and corrections given by different algorithms are shown in Table~\ref{tab:perspective_example}. The correction accuracy achieved by different approaches is reported in Table~\ref{tab:acc}. 

We divided toxic sentences into $5$ bins according to their original toxicities. In Fig.~\ref{fig:perpective_plot}, we report the average toxicity of the original, revised and corrected sentences in each bin. We see that the malicious errors drastically reduce the sentences' toxicities. The original average toxicity in the first bin is 0.55, whereas the revised toxicity is only 0.33 (a drop by $40\%$). This shows that  toxicity detection is very sensitive to spelling errors.

From the same figure we see that our proposed error correction  results in  toxicity scores closer to the original ones when  compared with the  baselines, validating the effectiveness of our approach. We note that in some cases our approach might result in a slightly higher toxicity score than the original one,  because of the pre-existing misspellings in the original sentences. For example, original sentences ``you are a \emph{fagget}'', ``\emph{fukkin} goof's track record'' and ``I suspect that closedmouth is \emph{g*y}'' have crude misspellings, which are used as Internet slang. Our approach  corrects these pre-existing misspellings in addition to the errors we add later, resulting in higher toxicity scores of corrected sentences. We perform corrections of all spelling errors, recovering more toxic words than we added maliciously.

\subsection{Spam Detection}
\label{sec:exp_spam}
\begin{table*}[htbp!]
\centering
\caption{Correction examples on spam data}
\label{tab:spam_example}
\resizebox{\textwidth}{!}{
\begin{tabular}{|l|l|l|l|}
\hline 
Original sentence & \begin{tabular}[c]{@{}l@{}}it really be a \textbf{great} oppotunity to make relatively \\ easy \textbf{money} , with little \textbf{cost} to you .\end{tabular} & \begin{tabular}[c]{@{}l@{}}we have quit our \textbf{jobs} , and will soon buy a home on \\ the beach and live off the interest on our \textbf{money} .\end{tabular} \\ \hline
Revised sentence & \begin{tabular}[c]{@{}l@{}}it really be a \textbf{grfat} opportunity to make relatively \\ easy \textbf{fmoney}, with little \textbf{cosgt} to you.\end{tabular} & \begin{tabular}[c]{@{}l@{}}we have quit our \textbf{tobs}, and will soon buy a home on \\ the beach and live off the interest on our \textbf{jmoney}.\end{tabular} \\ \hline
Our correction & \begin{tabular}[c]{@{}l@{}}it really be a \textbf{great} opportunity to make relatively \\ easy \textbf{money}, with little \textbf{cost} to you.\end{tabular} & \begin{tabular}[c]{@{}l@{}}we have quit our \textbf{jobs}, and will soon buy a home on \\the beach and live off the interest on our \textbf{money}.\end{tabular} \\ \hline
PyEnchant & \begin{tabular}[c]{@{}l@{}}it really be a \textbf{graft} opportunity to make relatively \\ easy \textbf{fmoney}, with little \textbf{cost} to you.\end{tabular} & \begin{tabular}[c]{@{}l@{}}we have quit our \textbf{bots}, and will soon buy a home on \\the beach and live off the interest on our \textbf{money}.\end{tabular} \\ \hline
Ekphrasis & \begin{tabular}[c]{@{}l@{}}it really be a \textbf{great} opportunity to make relatively \\ easy \textbf{money}, with little \textbf{cost} to you.\end{tabular} & \begin{tabular}[c]{@{}l@{}}we have quit our \textbf{tobs}, and will soon buy a home on \\ the beach and live off the interest on our \textbf{money}.\end{tabular} \\ \hline
Google & \begin{tabular}[c]{@{}l@{}}it really be a \textbf{great} opportunity to make relatively \\ easy \textbf{fmoney}, with little \textbf{cosgt} to you.\end{tabular} & \begin{tabular}[c]{@{}l@{}}we have quit our \textbf{jobs}, and will soon buy a home on \\the beach and live off the interest on our \textbf{jmoney}.\end{tabular} \\ \hline
\end{tabular}}
\end{table*}

\begin{figure}[htbp!]
\centering
\includegraphics[width=0.4\textwidth]{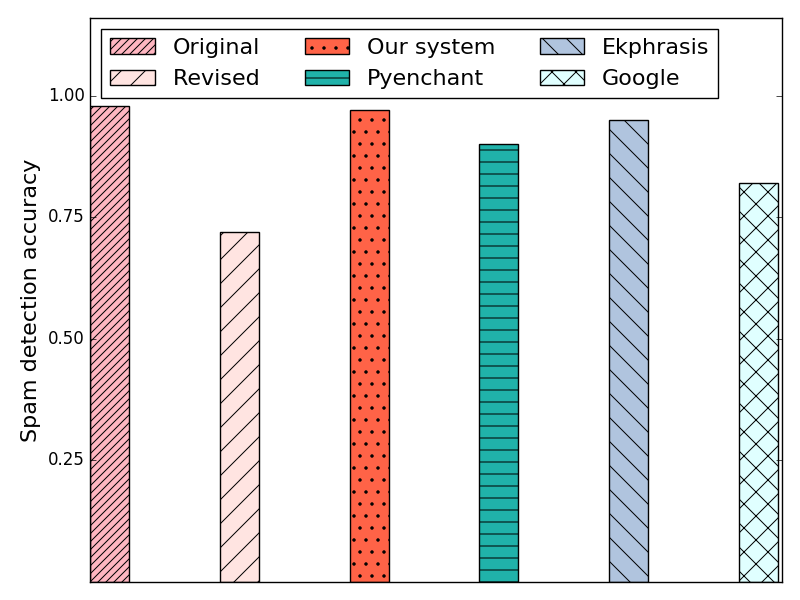}
\caption{Spam detection accuracy on test emails}
\label{fig:spam_plot}
\end{figure}

%Three approaches are applied to correct malicious errors in test spam data.
We next evaluate our spelling correction on the spam data. Synthetic malicious errors were added to spam-indicative words in spam mails based on our misspelling generation mechanism.

Some example spams are shown in Table~\ref{tab:spam_example}. Sensitive words such as ``money'' which are indicative of spams are highlighted, and the corrections given by different approaches are shown. Table \ref{tab:acc} shows the spelling correction accuracy achieved by our algorithm and the baselines for the spam data.

Fig.~\ref{fig:spam_plot} shows the spam detection accuracy on the original, revised and corrected test emails. Again we see that malicious errors resulted in a large accuracy drop,  the accuracy was restored after spelling correction, and the increase in accuracy using our approach was the maximal among those compared.

\subsection{Real Misspellings in Tweet Hate Speech Detection}
\label{sec:exp_cyberbullying}
\begin{table}[htbp!]
\centering
\caption{Example of tweets and hate speech categories}
\label{tab:tweet_example}
\resizebox{\linewidth}{!}{
\begin{tabular}{|c|l|l|}
\hline
\begin{tabular}[c]{@{}c@{}}Category\end{tabular} & \begin{tabular}[c]{@{}c@{}} Example sentence\end{tabular} \\ \hline
Racism & \begin{tabular}[l]{@{}l@{}}\#isis \#islam pc puzzle: converting to a religion \\ of peace leading to violence? http://t.co/tbjusaemuh \end{tabular} \\ \hline
Sexism & \begin{tabular}[l]{@{}l@{}}Real question is do feminist liberal bigots \\understand that different rules for men/women \\is sexism\end{tabular} \\ \hline
Neither & The mother and son team are sooooo nice !!! \\ \hline
\end{tabular}}
\end{table}

We have shown that synthetic misspellings are able to deceive the toxicity detector and the spam detector, and reported the performance of spell checkers on the synthetic data. We also experimented with data containing real spelling errors collected from Twitter, where instances of hate speech contain user-generated spelling errors. The correction performance of spell checkers are now compared on the real misspellings.

%\textbf{!!!Change example table based on the content!!!}

% \usepackage{multirow}
\begin{table*}[htbp!]
\centering
\caption{Hate speech detection results with different spelling corrections} %\textcolor{red}{Check Google's recall for racist and F1 for sexist. Which ones should be in bold?}}
\label{tab:cyberbullying_detection}
\begin{tabular}{|c|c|c|c|c|c|c|}
\hline
Category & Metric & Original & Our system & PyEnchant & Ekphrasis & Google \\ \hline
\multirow{3}{*}{racist} & Precision & 0.630 & \textbf{0.640} & 0.630 & 0.630 & 0.569 \\ \cline{2-7} 
 & Recall & 0.617 & 0.681 & 0.617 & 0.617 & \textbf{0.702} \\ \cline{2-7} 
 & F1 score & 0.623 & \textbf{0.660} & 0.623 & 0.623 & 0.628 \\ \hline
\multirow{3}{*}{sexist} & Precision & 0.641 & 0.630 & 0.629 & 0.629 & \textbf{0.649} \\ \cline{2-7} 
 & Recall & 0.775 & 0.767 & \textbf{0.790} & 0.790 & 0.759 \\ \cline{2-7} 
 & F1 score & 0.701 & 0.692 & \textbf{0.701} & \textbf{0.701} & 0.700 \\ \hline
\multirow{3}{*}{\begin{tabular}[c]{@{}c@{}}Macro average over \\ all categories\end{tabular}} & Precision & 0.721 & \textbf{0.721} & 0.718 & 0.718 & 0.703 \\ \cline{2-7} 
 & Recall & 0.741 & 0.757 & 0.743 & 0.743 & \textbf{0.759} \\ \cline{2-7} 
 & F1 score & 0.727 & \textbf{0.737} & 0.727 & 0.727 & 0.727 \\ \hline
\end{tabular}
\end{table*}

\textbf{Tweet normalization}. Some example tweets of racist and sexist nature are shown in Table~\ref
{tab:tweet_example}. Tweets are notoriously noisy and unstructured given the frequent occurrences of hashtags, URLs, reserved words and emojis. These non-standard tokens greatly increase the vocabulary size of tweets while also injecting noise to classification tasks. We use Tweet Preprocessor, a tweet preprocessing tool which can replace aforementioned tokens with a special set of tokens. For example, one tweet is ``\#isis \#islam pc puzzle: converting to a religion of peace leading to violence ?,http://t.co/tbjusaemuh http://t.co/g4xoh...'', which becomes ``\$HASHTAG\$ \$HASHTAG\$ pc puzzle: converting to a religion of peace leading to violence? \$URL\$ \$URL\$...'' after preprocessing. This preprocessing stage can clean texts without dealing with misspellings. Tweets are preprocessed right before they are fed into the neural network.

\textbf{Hate speech detection}. The state-of-the-art system for hate speech detection is a Bidirectional Long Short Term Memory (BLSTM) network with attention mechanism \cite{agrawal2018deep}. The model takes a sequence of words in a tweet, obtains word embeddings in the embedding layer, and passes them to bidirectional recurrent layers to generate a dense representation of the input tweet. The feedforward layer takes the tweet vector and predicts its probability distribution over all three classes. The class with the highest likelihood is chosen as the category of the input tweet. In our experiment, we use a BLSTM model for the hate speech detection task. 

\textbf{Vocabulary construction}. Firstly we build a vocabulary list using both a standard dictionary and the frequent words (frequency higher than 5) in the training data. This list will serve as a reference for spelling correction; words outside the list will be taken as spelling errors and replaced with legal words from the vocabulary. The reason for collecting words from the training data is to include the Internet slangs in tweets which may not exist in the standard dictionary. Some examples are ``tmr'' (for \emph{tomorrow}), ``fwd'' (for \emph{forward}) and ``lol'' (for \emph{laughing out loudly}). Some previous works proposed to replace these slang terms with their standard forms \cite{gupta2017tweet,modupe2017semi}, which require either expert knowledge or human annotations. We argue that because these Internet slangs change rapidly we should understand them in a data-driven manner instead of standardizing them based on human knowledge. Since the representation of these slangs and their use in hate speech  will be learned by the neural network from the train data, we add these slangs to our vocabulary as legal words.

\textbf{Spelling correction}. Misspellings are another source of text noise which cannot be handled in the tweet normalization stage. Some misspellings are user-created to deceive the online detection system. For example, the swear word ``fucking'' has a lot of variants in tweets such as ``fckn'', ``f*ckin'', ``f**king'', ``fckin'', ``fuckin'', ``fking'',``fkin'' and ``fkn''. When a new variant of a swear word arises in the test data, the model takes it as an out-of-vocabulary word, and is unable to match it with any learned pattern. The purpose of spelling correction is to map these new variants to words that are known to the model. As discussed in Section~\ref{sec:model}, we enumerate the candidates of a misspelled word, and choose the candidate which best fits with the context as its correction.

\textbf{Results}. We  do a random 80:20 train-test split of the Twitter dataset. A  detection system was trained on the normalized train data (without spelling correction) using the state-of-the-art BLSTM model. There were $3218$ test tweets, $571$ of which had misspellings. 
We applied the different spelling correction approaches to these $571$ tweets, and the corrected tweets were then cleaned as described in the tweet normalization stage. Processed tweets were input to the trained detection system.
We compared the hate speech detection results on the $571$ test tweets to evaluate the effect of spelling correction  on this downstream application. As shown in Table~\ref{tab:cyberbullying_detection}, We report results on the original test data, and also on the test data which are corrected by our approach, PyEnchant, Ekphrasis, and google spell checker.
We report precision, recall and F1-score of racist and sexist classification respectively and the macro-averages (to evaluate the overall performance).%, we also report macro-averages of precision, recall and macro F1 score, which are the average of corresponding metrics for each class.

The best performance of each metric is highlighted in the table. Compared with the classification performance on the test data with spelling errors for the racist category, our approach improves the precision  by $1$\%, the recall by $6.4$\%, and the F1 score by $3.7$\% absolute  points. Both PyEnchant and Ekphrasis spell checkers improve the recall of sexist category, but decrease the precision, so their corrected forms achieve  F1 scores similar to the original test data. Google spell checker also gives similar F1 score on both racist and sexist classes. Our approach outperforms the other baselines in terms of F1 score for the racist class and the macro F1 score.

% correct example ``the stupidist arguments are the ones rationalizing the barbarity of islam.''

For sexism-related tweets, our approach does not improve the results compared to  the original test data. Taking a closer look at the nature of the sexist tweets we notice that they often contain some abbreviations which might be taken as misspellings, and that their contextual information is insufficient to decide the appropriate corrections. For example, in the sexist tweet ``I'm sorry but if you watch women ufc fights kys''.  Our approach replaces \emph{kys} with \emph{keys}, and the trained neural network misclassified the corrected tweet. Another sexist-related tweet is ``these nsw promo girls think way too highly of themselves'', where \emph{nsw} is incorrectly replaced with \emph{new} by our approach.

\section{Conclusion}
In this study, we showed how malicious spelling errors can deceive profanity- and spam detectors. To deal with these malicious misspellings, we proposed a context-sensitive spelling corrector based on word embeddings. Our spell checker is light-weight, unsupervised and can be easily incorporated into downstream applications. It achieved a  favorable spelling correction performance when compared with general purpose spell-checking tools such as PyEnchant, Ekphrasis and Google spell checkers on both synthetic and real misspellings from different datasets.

\bibliographystyle{acl_natbib}
\bibliography{sample-bibliography}

\end{document}